\documentclass[runningheads]{llncs}

\usepackage{eccv}

\usepackage{eccvabbrv}

\usepackage[utf8]{inputenc} 
\usepackage[T1]{fontenc}    
\usepackage{graphicx}
\usepackage{booktabs}       
\usepackage{xcolor}         
\usepackage{url}            
\usepackage{amsmath}
\usepackage{amsfonts}       
\usepackage{nicefrac}       
\usepackage{microtype}      
\usepackage{multirow}
\usepackage{tabularx}
\usepackage{makecell}
\usepackage{array}
\usepackage[normalem]{ulem}
\usepackage{enumitem}
\usepackage{bm}

\usepackage[accsupp]{axessibility}  

\usepackage{hyperref}

\usepackage{orcidlink}

\begin{document}

\title{DSH-Bench: A Difficulty- and Scenario-Aware Benchmark with Hierarchical Subject \\ Taxonomy for Subject-Driven Text-to-Image Generation} 

\titlerunning{DSH-Bench: A comprehensive benchmark for Subject-Driven T2I}

\author{Zhenyu Hu\inst{1}\thanks{Equal contribution. $^\dagger$ Corresponding author. $^\ddagger$ Project leader.} \and
Qing Wang\inst{1}$^\star$ \and
Te Cao\inst{1}$^\star$ \and
Luo Liao\inst{1}$^\star$ \and
Longfei Lu\inst{1} \and
Liqun Liu\inst{1}$^\dagger$ \and
Shuang Li\inst{1} \and
Hang Chen\inst{1} \and
Mengge Xue\inst{1}$^\ddagger$ \and
Yuan Chen\inst{1} \and
Chao Deng\inst{1} \and
Peng Shu\inst{1} \and
Huan Yu\inst{1} \and
Jie Jiang\inst{1}}

\authorrunning{Z. Hu et al.}

\institute{Tencent, China \\
\email{\{mapleshu, rosaliecao, graceqwang, berryxue, liqunliu\}@tencent.com}}

\maketitle

\begin{abstract}

Significant progress has been achieved in subject-driven text-to-image (T2I) generation, which aims to synthesize new images depicting target subjects according to user instructions. However, evaluating these models remains a significant challenge. Existing benchmarks exhibit critical limitations: 1) insufficient diversity and comprehensiveness in subject images, 2) inadequate granularity in assessing model performance across different subject difficulty levels and prompt scenarios, and 3) a profound lack of actionable insights and diagnostic guidance for subsequent model refinement. To address these limitations, we propose DSH-Bench, a comprehensive benchmark that enables systematic multi-perspective analysis of subject-driven T2I models through four principal innovations: 1) a hierarchical taxonomy sampling mechanism ensuring comprehensive subject representation across 58 fine-grained categories, 2) an innovative classification scheme categorizing both subject difficulty level and prompt scenario for granular capability assessment, 3) a novel Subject Identity Consistency Score (SICS) metric demonstrating a 9.4\% higher correlation with human evaluation compared to existing measures in quantifying subject preservation, and 4) a comprehensive set of diagnostic insights derived from the benchmark, offering critical guidance for optimizing future model training paradigms and data construction strategies. Through an extensive empirical evaluation of 19 leading models, DSH-Bench uncovers previously obscured limitations in current approaches, establishing concrete directions for future research and development.
\keywords{Single Subject-driven \and Text-to-image Generation \and Dataset
and Benchmark \and Subject Identity Consistency}

\end{abstract}

\begin{figure}[t]
\centering
\includegraphics[width=1.0\textwidth]{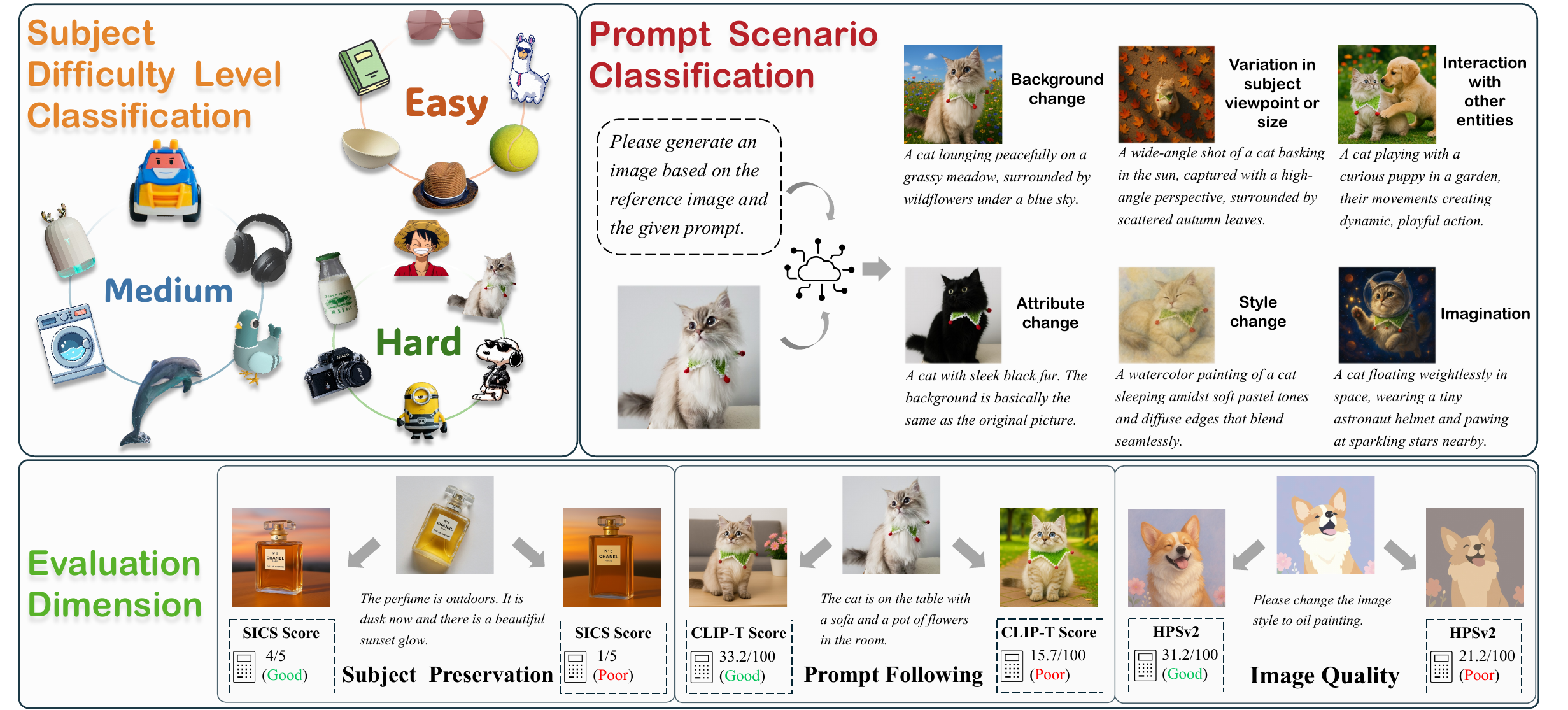}
\caption{\textbf{Overview of DSH-Bench}. We curate a diverse dataset of subject images and categorize them into three difficulty levels---\textbf{easy}, \textbf{medium}, and \textbf{hard}---based on the complexity of preserving subject details. Leveraging GPT-4o's capabilities, we systematically generate contextually appropriate prompts for various scenarios. The generated images are then rigorously evaluated across three key dimensions: \textbf{Subject Preservation}, \textbf{Prompt Following}, and \textbf{Image Quality}.}
\label{fig:overview}
\end{figure}

\section{Introduction}

Subject-driven text-to-image (T2I) generation, which synthesizes novel scenes conditioned on specific reference images and textual prompts, has emerged as a pivotal research frontier, propelled by rapid advancements in large-scale T2I diffusion models~\cite{ding2021cogview, gafni2022make, saharia2022photorealistic, rombach2022high, balaji2022ediff, chang2023muse, kang2023scaling, dong2024dreamllm}. In subject-driven T2I generation, aside from image quality, two other fundamental criteria must be satisfied: Subject Preservation and Prompt Following. Subject Preservation requires that the generated image maintain the details of the reference subject. Prompt Following demands that the generated image consistently reflects the content in the prompt. For example, a user might request an image of "his dog traveling around the world". In this scenario, the generated image must depict a dog identical to the reference image while illustrating the act of traveling as described.

\begin{figure*}[t]
\centering
\includegraphics[width=1.0\textwidth]{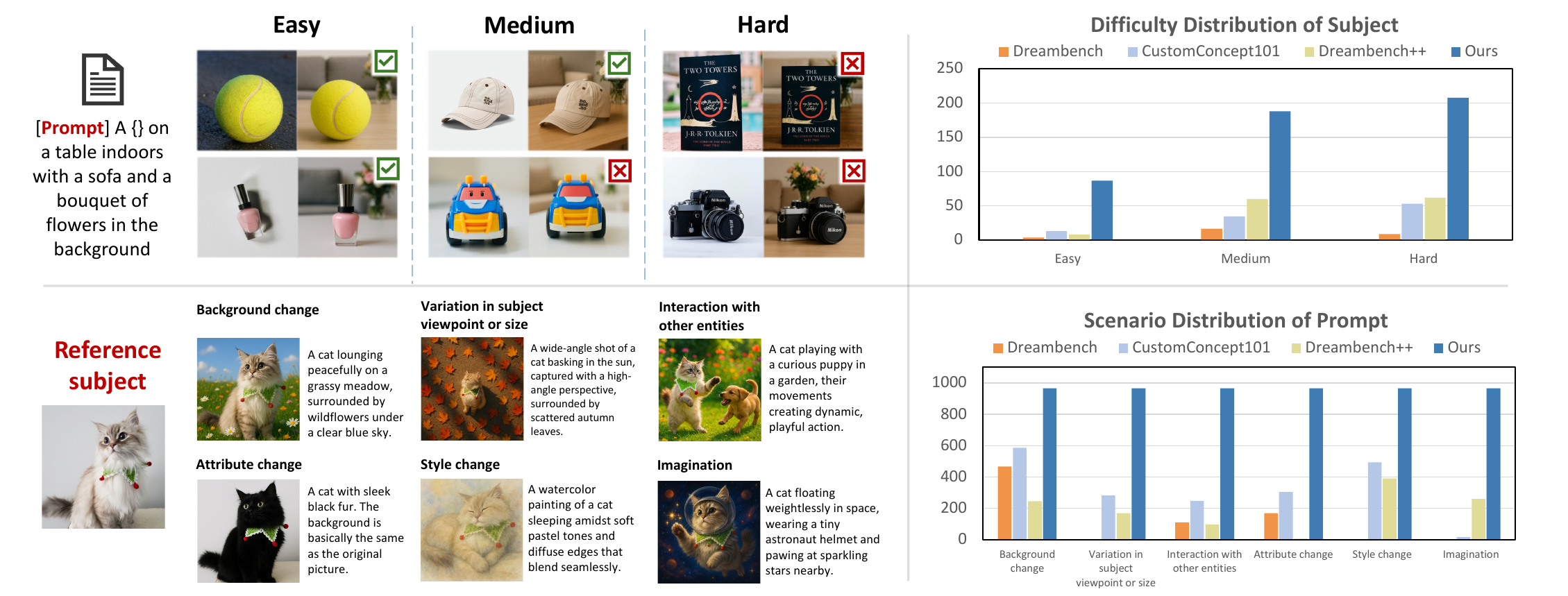}
\caption{Qualitative comparison under different difficulty levels and scenarios.}
\label{fig:difficulty scene diff}
\end{figure*}

Although significant progress has been made in subject-driven T2I generation in recent years~\cite{ruiz2023dreambooth, gal2022textual, kumari2023multi, wang2024instantstyle, li2023blip, ye2023ip, gal2023encoder, wei2023elite, hu2024instruct, qiu2023controlling}, the community still lacks a standardized protocol to comprehensively and effectively evaluate their true capabilities. An optimal benchmark should not only ensure unbiased, human-aligned evaluation but also provide granular diagnostics to guide research. However, current benchmarks~\cite{ruiz2023dreambooth, kumari2023multi, chen2023subject, wang2024ms, peng2024DreamBench} are limited by insufficient diversity and comprehensiveness in subject image collection, which restricts the thoroughness of model evaluation. Crucially, they fail to disentangle the inherent difficulty of the reference subjects from the complexity of the prompt scenarios. As shown in \cref{fig:difficulty scene diff}, our empirical analysis reveals a significant performance variance contingent on these factors: models that effortlessly reconstruct simple geometries (e.g., a tennis ball) frequently fail to preserve the intricate structural details of complex artifacts (e.g., a camera). This observation highlights a critical blind spot in current evaluations: the necessity of stratifying test cases by the subject difficulty and prompt scenario. Moreover, subject-driven T2I generation can be broadly categorized into single-subject and multi-subject paradigms depending on the number of reference subjects. While the single-subject task can be conceptually considered a special case of the multi-subject scenario, the two paradigms impose fundamentally different demands on model capabilities. Regarding the scope of evaluation, we posit that the single-subject paradigm is the cornerstone of subject-driven generation. While multi-subject generation introduces interactional complexities, empirical evidence suggests that models still struggle to achieve saturated performance even in single-subject tasks. Consequently, establishing a rigorous benchmark for the single-subject scenario is a prerequisite for any reliable multi-subject assessment.

\begin{figure*}[t]
\centering
\includegraphics[width=1.0\textwidth]{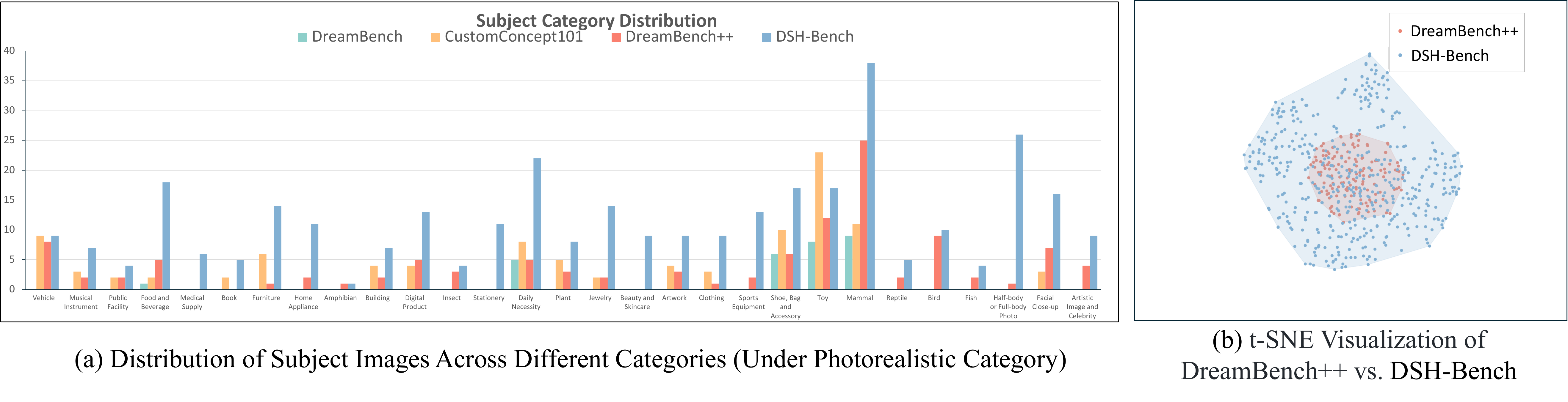}
\caption{\textbf{Distribution of subject images.} (a) Category-wise image distribution for our benchmark versus prior benchmarks. (b) t-SNE comparison of images between DSH-Bench and DreamBench++.}
\label{fig:image_category_dist}
\end{figure*}

To address the aforementioned limitations, we introduce DSH-Bench, a novel and comprehensive benchmark designed to provide multi-dimensional evaluations. An overview of DSH-Bench is illustrated in \cref{fig:overview}. DSH-Bench offers three distinct advantages:






1.\textbf{\textit{The diversity of subject images in DSH-Bench is substantially greater}}\quad To effectively mitigate potential evaluation bias stemming from limited subject diversity, we construct a rigorous hierarchical taxonomy for image collection. Specifically, we incorporate established ontologies from COCO~\cite{lin2014microsoft} and ImageNet~\cite{deng2009imagenet} into our taxonomy design. As illustrated in \cref{fig:image_category_dist}(a), the widely utilized DreamBench comprises merely 6 categories and 30 subjects. In sharp contrast, our benchmark significantly scales up the dataset to 58 distinct categories and 459 unique subjects---representing a substantial increase of \textbf{9}$\times$ and \textbf{15}$\times$, respectively. While DreamBench++~\cite{peng2024DreamBench} offers 150 subjects, its diversity remains constrained by a limited collection scope. Notably, \textbf{33\%} of our categories are entirely absent from the DreamBench++ distribution. Consequently, benefiting from DSH-Bench's superior subject diversity, we facilitate a far more comprehensive and robust evaluation of generative models.

2. \textbf{\textit{An innovative classification scheme for subject difficulty and prompt scenarios}}\quad As illustrated in \cref{fig:difficulty scene diff}, model performance exhibits significant variance across different samples, underscoring the critical necessity for a dual classification of both subject images and prompts. While DreamBench++ attempts to categorize prompts based on perceived difficulty, the underlying criteria remain ambiguous and lack a systematic definition. Furthermore, it neglects to analyze the inherent difficulty levels associated with distinct subjects. To address these limitations, we systematically stratify subjects into three difficulty tiers (easy, medium, hard) based on the complexity of preserving visual appearance, and classify prompts into six distinct scenarios (background change, variation in subject viewpoint or size, interaction with other entities, attribute change, style transfer, and imagination). Thus, our approach enables a more comprehensive and granular diagnosis of the challenges faced by current models.



3. \textbf{\textit{A human-aligned and more efficient metric for subject preservation}}\quad DreamBench++ replaces CLIP~\cite{radford2021learning} and DINO~\cite{caron2021emerging} with GPT-4o~\cite{openai2024gpt4o} for evaluation, resulting in improved alignment with human evaluation. However, our benchmark reveals that per-model evaluation under this paradigm requires approximately 20,000 API calls to GPT-4o, incurring prohibitive computational costs exceeding \$400 for each evaluation. To address the limitation, we introduce \textbf{Subject Identity Consistency Score} (SICS), which innovatively focuses on subject-level consistency rather than merely relying on embedding comparisons. Firstly, five annotators label a training dataset containing 5,000 image-text pairs, focusing on subject preservation evaluation. We then fine-tune Qwen2.5-VL-7B~\cite{bai2025qwen25vltechnicalreport} on this dataset, which leads the model to focus on core visual attributes rather than high-level semantics. Finally, we use Kendall's $\tau$ value to quantify the alignment between model outputs and human evaluation. Experimental results demonstrate that SICS achieves a statistically significant improvement, outperforming Dreambench++ by 9.4\% in human evaluation correlation metrics.

\textbf{Takeaways}\quad We present some insightful findings from evaluating 19 methods: i) Our evaluation reveals that no single method demonstrates consistently robust performance across all categories. Therefore, implementing hierarchical taxonomy sampling of subject images is critical for mitigating potential evaluation biases. ii) All methods exhibit degraded performance on hard subject images. It is crucial to enhance models’ ability to encode and reconstruct complex subject details more effectively in future research. iii) The subject-driven T2I model's capability for different prompt scenarios is not robust. Future research on subject-driven T2I generation should focus on optimizing for adaptation to a variety of prompt scenarios.

In summary, our contributions are as follows: 1) We employ a hierarchical taxonomy in image collection to ensure both the diversity and comprehensiveness of subject images. 2) We propose an innovative classification scheme to categorize subject difficulty levels and prompt scenarios. This scheme enables us to obtain valuable insights. 3) We propose a human-aligned metric to evaluate subject preservation, which offers greater efficiency compared to DreamBench++. We open-source DSH-Bench, including subject images, prompts and code, at \url{https://github.com/sunriverhzy/DSH-Bench}.


\section{Related Work}

\subsection{Subject-Driven Text-to-Image Generation}
\noindent  In recent years, subject-driven T2I generation has attracted significant research attention~\cite{ruiz2023dreambooth, gal2022textual, kumari2023multi, wang2024instantstyle, li2023blip, gal2023encoder, gal2023an, wei2023elite, hu2024instruct, qiu2023controlling}. Within the context of diffusion models, optimization-based model~\cite{
voynov2023pextendedtextualconditioning, 
liu2023cones2customizableimage, 
hua2023dreamtunersingleimagesubjectdriven,
hao2023vico} enables subject-driven generation by introducing lightweight parameters and performs parameter-efficient fine-tuning for each subject. In contrast, the encoder-based methods~\cite{
shi2023instantboothpersonalizedtexttoimagegeneration, ma2024subjectdiffusionopendomainpersonalizedtexttoimage, 
chen2023dreamidentityimprovededitabilityefficient, 
li2023photomakercustomizingrealistichuman,
le2024diffusiongenerate,
rowles2024ipadapterinstructresolvingambiguityimagebased,
zeng2024jedijointimagediffusionmodels,
hu2024instructimagenimagegenerationmultimodal,
huang2025flexip,
xiong2025groundingbooth,
patashnik2025nestedattention,
wu2025less,
huang2025wegen,
he2025anystory} leverage additional image encoders and network layers to encode the reference image of the subject. IP-Adapter~\cite{ye2023ip} introduces cross-attention through an additional image encoder to incorporate control signals. Furthermore, SSR-Encoder~\cite{zhang2024ssr} enhances identity preservation without necessitating further fine-tuning when introducing new concepts. The Diffusion Transformers~\cite{peebles2023scalablediffusionmodelstransformers,podell2023sdxlimprovinglatentdiffusion, rombach2022highresolutionimagesynthesislatent} uses transformer as a denoising network to iteratively refine noisy image tokens. Based on these foundation models, approaches like OminiControl~\cite{tan2024ominicontrol} and UNO ~\cite{wu2025less} explore the inherent image reference capabilities of transformers.


\noindent \subsection{Subject-Driven T2I Generation Benchmark}
\noindent Evaluation for subject-driven T2I generation involves a variety of metrics focusing on different aspects. For image quality, several notable studies~\cite{xu2023imagereward, kirstain2023pick, wu2023human, alaluf2023neural, wang2025unified} have conducted. For subject preservation evaluation, learning-based metrics~\cite{NIPS2016_371bce7d, 8578166, 8578292} compute the distances between image features extracted by deep neural networks. Image embeddings from large vision models like CLIP~\cite{radford2021learning}, DINO~\cite{caron2021emerging} and image-retrieval score~\cite{liu2021image} have been utilized. To better match human perception, DreamSim~\cite{fu2023dreamsim} focuses on foreground objects when evaluating image similarity. In terms of semantic consistency, the CLIP score is frequently used. More recently, RefVNLI~\cite{slobodkin2025refvnli} jointly evaluates textual alignment and subject preservation within a single prediction. Dreambench~\cite{ruiz2023dreambooth} is the first benchmark established for subject-driven T2I tasks. However, Dreambench is limited in the diversity of subjects and prompt scenarios. DreamBench-v2~\cite{chen2023subject} extends the evaluation by introducing 220 novel prompts. CustomConcept101~\cite{kumari2023multi} lacks of non-photorealistic subject images. Although DreamBench++~\cite{peng2024DreamBench} increases to 150 subject images, it can not provide a systematic categorization of subjects and prompts, which makes it difficult to derive meaningful insights from the evaluation results. More recent efforts such as MMIG-Bench~\cite{hua2026mmig} further broaden evaluation toward multi-modal image generation across diverse tasks. To bridge this gap, we introduce DSH-Bench, which enables a more comprehensive evaluation of models for subject-driven T2I generation and provids deeper insights.

\section{DSH-Bench}

This section provides an overview of the primary components of DSH-Bench. \cref{sec:benchmark dataset construction} outlines the data construction process. \cref{sec:evaluation dimension} introduces the definitions and evaluation methods for three evaluation dimensions. \textit{A detailed explanation is available in the supplementary materials.}

\begin{figure*}[t]
\centering
\includegraphics[width=1.0\textwidth]{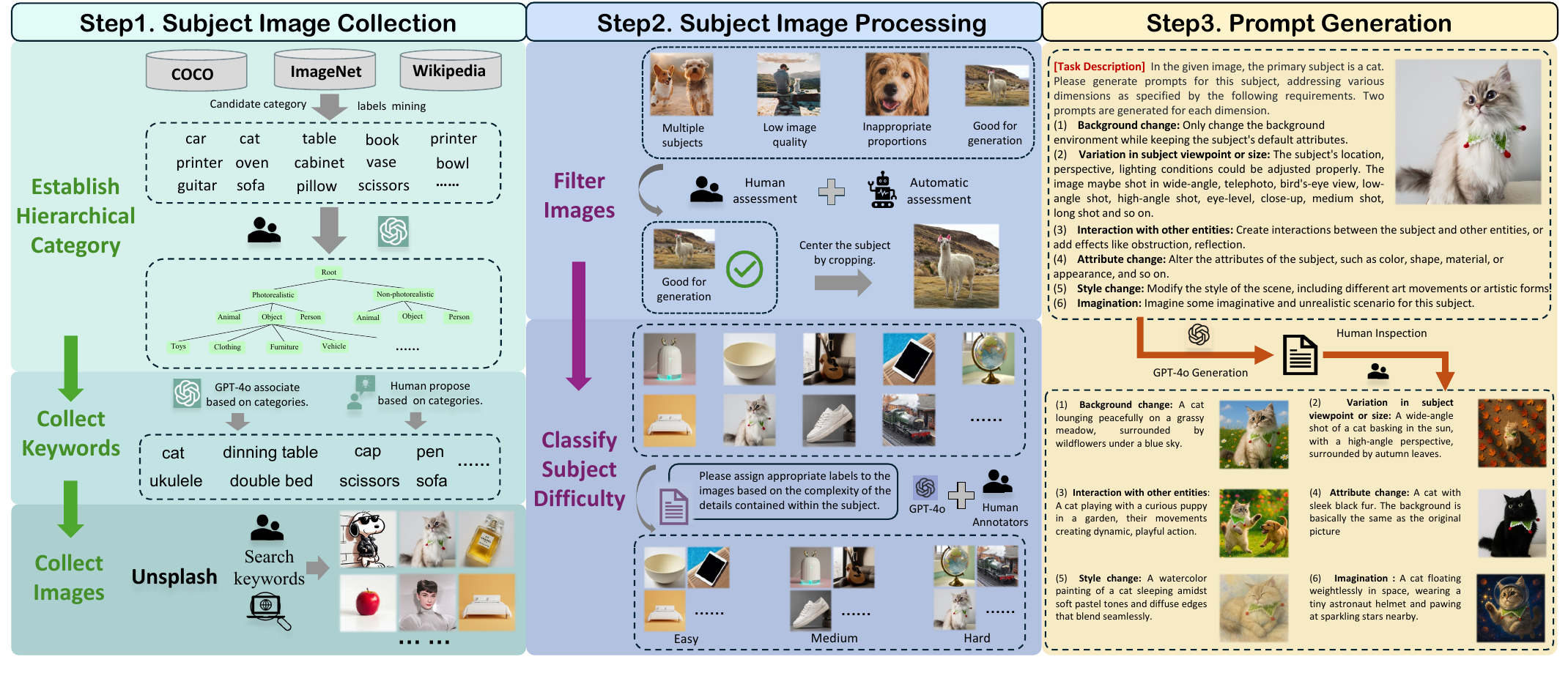}
\caption{\textbf{Dataset construction process of DSH-Bench.} We construct a hierarchical taxonomy to obtain a comprehensive set of keywords. Then we collect web images using these keywords. After performing both manual review and automated filtering of the images, we classify the difficulty of subject images and use GPT-4o to generate prompts for each subject image.}
\label{fig:data_construct}
\end{figure*}

\subsection{Benchmark Dataset Construction}
\label{sec:benchmark dataset construction}
\subsubsection{Subject Image Collection} \quad \textit{\textbf{1) Hierarchical Taxonomy Establishment}}\quad As illustrated in \cref{fig:data_construct}, we construct a rigorous three-tier taxonomy. The first level distinguishes between \textit{Photorealistic} and \textit{Non-photorealistic} domains, with strictly aligned subcategories to ensure cross-domain consistency. For the second level, we refine the coarse "Living" category from prior benchmarks~\cite{ruiz2023dreambooth, peng2024DreamBench} into distinct \textit{Humans} and \textit{Animals}, acknowledging the unique demand for facial fidelity in human subjects versus the high variance in animals. We explicitly exclude abstract "Style" categories~\cite{kumari2023multi} to focus strictly on tangible entity customization. At the third level, to balance granularity with generality, we compiled candidate labels from COCO and ImageNet, utilizing GPT-4o to consolidate them into 58 fine-grained categories. Notably, we stratify the \textit{Human} category into celebrities, facial close-ups, and full-body shots, allowing us to disentangle foundation model bias (e.g., celebrity overfitting) from structural reconstruction capabilities. \quad \textit{\textbf{2) Keyword Collection \& Internet Image Collection}}\quad In DreamBench++, keywords collection relies on GPT-4o and human input. The approach does not adequately ensure the diversity of the obtained keywords, potentially introducing bias during the image collection process. In contrast, DSH-Bench derives keywords from a hierarchical taxonomy. For each third-level category, we use GPT-4o to generate associated keywords, which are further supplemented by humans. All keywords are then consolidated and deduplicated, resulting in a final set of \textbf{400} unique keywords—surpassing DreamBench++'s 300. The specific keywords are provided in the Supplementary material (Sec. B). Given a set of selected keywords, we retrieve images from Unsplash~\cite{unsplash}. \textit{Each image’s copyright status has been verified for academic suitability}.

 
\subsubsection{Subject Image Processing} \quad \textit{\textbf{1) Image Filtering}} \quad To filter unsuitable images, we use aesthetic score~\cite{xu2024visionreward} and SAM~\cite{kirillov2023segment} to filter images with low image quality and inappropriate proportions of subject regions. The curated images are subsequently cropped to centralize the reference subject. \quad \textit{\textbf{2) Subject Difficulty Level Classification}}\quad As illustrated in \cref{fig:difficulty scene diff}, the model's performance varies considerably across different samples. To derive meaningful insights, we classify the subject images according to the difficulty level that the model experiences in preserving details of the reference subject. We define three subject difficulty levels, including (1)\textbf{ Easy:} Subjects characterized by minimal surface complexity and homogeneous textural properties, exemplified by smooth-surfaced objects such as a ceramic mug with uniform coloration. These cases present negligible challenges for detail preservation due to their structural regularity. (2)\textbf{ Medium:} Subjects containing discernible high-frequency features while maintaining global structural coherence, such as cylindrical containers with legible typographic elements. These cases require intermediate detail preservation capabilities. (3)\textbf{ Hard:} Subjects exhibiting non-uniform texture distributions and multi-scale geometric details, typified by objects like book covers containing fine-grained calligraphic elements. Such cases expose model limitations in maintaining structural fidelity and textural granularity under complex topological constraints. We utilize GPT-4o to classify the subject images according to the aforementioned criteria. Subsequently, all images are reviewed by five human annotators to ensure accuracy and consistency. As these images are curated based on a meticulously constructed taxonomy, they possess the comprehensiveness and representativeness required for rigorous evaluation. Given the inference efficiency of current generative models, an overabundance of redundant images would substantially inflate the evaluation overhead. Ultimately, we obtain a total of \textbf{459} high-quality images. \quad \textit{\textbf{3) Validation of Difficulty Labels against Objective Image Properties}}\quad To verify that our difficulty labels reflect intrinsic image complexity rather than model-dependent heuristics, we randomly sample 50 images per difficulty level (5 independent runs) and compute three model-independent complexity metrics: Canny edge density, GLCM contrast, and JPEG bits/pixel (quality factor $Q=85$). As reported in \cref{tab:difficulty_metrics}, all three metrics increase monotonically from Easy to Medium to Hard, and each exhibits a statistically significant positive Spearman correlation with the difficulty level ($p<0.01$). This confirms that our difficulty labels correlate with objective, model-agnostic image properties, rather than relying on circular model-dependent definitions.

\begin{table}[tb]
\centering
\setlength{\tabcolsep}{1pt}
\caption{\textbf{Validation of difficulty labels against model-independent image-complexity metrics.} For each difficulty level, we randomly sample 50 images (5 independent runs) and report the mean $\pm$ std of Canny edge density, GLCM contrast, and JPEG bits/pixel ($Q=85$). All three metrics increase monotonically from Easy to Hard and correlate significantly with the difficulty level.}
\label{tab:difficulty_metrics}
\scalebox{1}{
\begin{tabular}{lccccc}
\toprule
Metric & Easy & Medium & Hard & Spearman $\rho$ & $p$-value \\
\midrule
Edge density    & $0.024 \pm 0.004$ & $0.048 \pm 0.006$ & $0.053 \pm 0.007$ & 0.367 & $3.2 \times 10^{-6}$ \\
GLCM contrast   & $3.71 \pm 0.20$   & $4.07 \pm 0.26$   & $5.28 \pm 0.43$   & 0.235 & $2.5 \times 10^{-3}$ \\
JPEG bits/pixel & $0.71 \pm 0.04$   & $0.91 \pm 0.06$   & $0.96 \pm 0.04$   & 0.260 & $1.3 \times 10^{-3}$ \\
\bottomrule
\end{tabular}
}
\end{table}



\subsubsection{Prompt Generation} \quad Although DreamBench++ categorizes prompts based on their perceived difficulty, it does not provide  empirical evidence to substantiate the criterion. To address this limitation, we organize the prompts according to specific application scenarios, dividing them into six categories, including (1)\textbf{ Background change (BC):} scenarios involving changes in background elements. (2)\textbf{ Variation in subject viewpoint or size (VS):} scenarios that entail changes in camera angle, which may include variations in subject size, lighting, or shadows. (3)\textbf{ Interaction with other entities (IE):} scenarios requiring complex interactions with additional entities, potentially resulting in occlusion and necessitating adherence to physical plausibility. (4)\textbf{ Attribute change (AC):} scenarios involving modifications to certain attributes of the subject, such as color or shape. (5)\textbf{ Style change (SC):} scenarios involving alterations in the artistic or visual style of the subject. (6)\textbf{ Imagination (IM):} scenarios where the target image depicts an imagined or fictional scene. We generate two prompts for each scenario. The specific instructions are depicted in \cref{fig:data_construct}. All prompts are reviewed by five human annotators to ensure they are ethical and free from defects. For the specific verification procedure, please refer to the Supplementary material (Sec. E.3). Finally, we obtain a total of \textbf{5,508} prompts. \cref{fig:difficulty scene diff} shows the distribution of subject difficulty levels and prompt scenarios. We visualize the t-SNE of images from our benchmark and DreamBench++ in \cref{fig:image_category_dist}, which indicate DSH-Bench achieves superior diversity.


\begin{figure*}[t]
\centering
\includegraphics[width=1.0\textwidth]{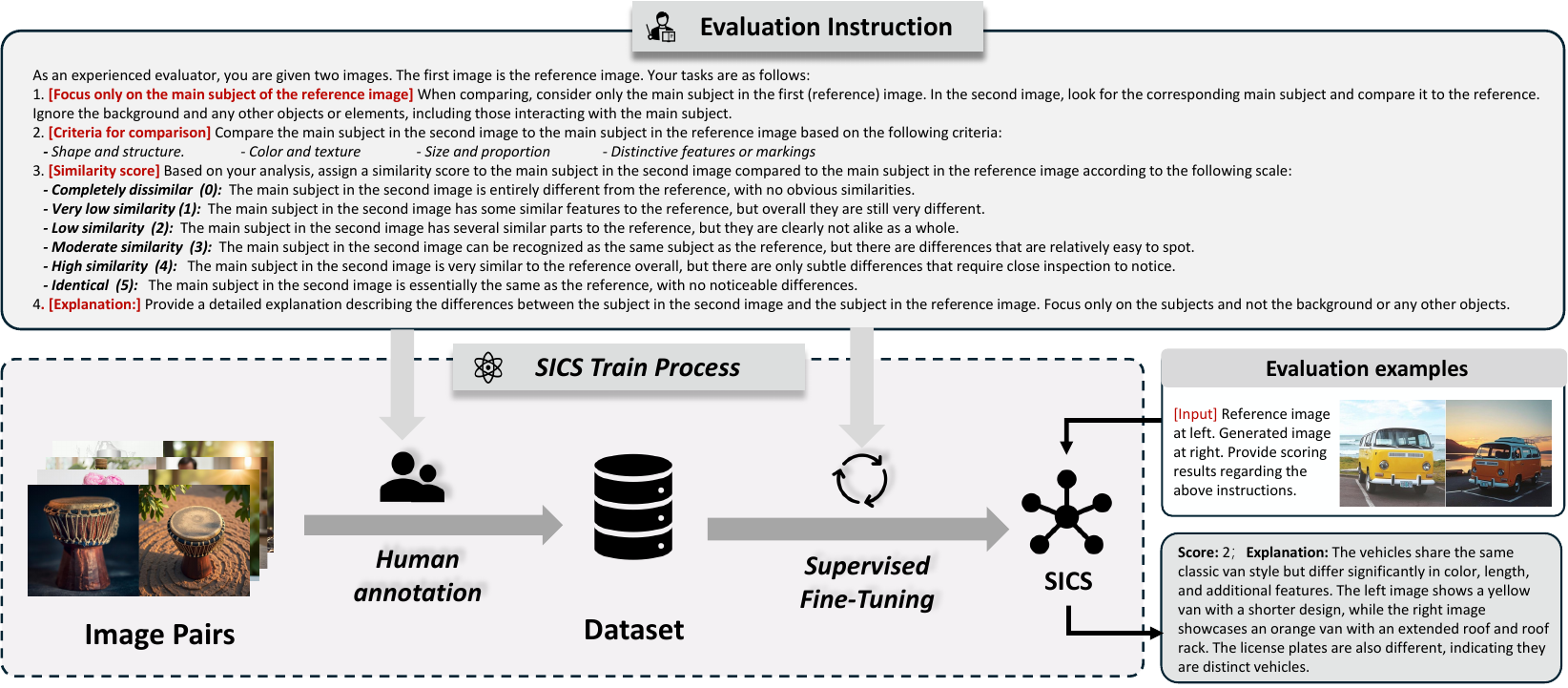}
\caption{\textbf{The training process of SICS.} We constructed and annotated a dataset specifically tailored for subject consistency determination, and subsequently trained models using this dataset.}
\label{fig:SCIS process}
\end{figure*}

\subsection{Evaluation Dimension}
\label{sec:evaluation dimension}

Previous notable works~\cite{ruiz2023dreambooth, gal2022textual, kumari2023multi, wang2024instantstyle} evaluate the performance of subject-driven T2I models from two perspectives: Subject Preservation and Prompt Following. RealCustom++~\cite{mao2024realcustom++} also uses ImageReward~\cite{xu2023imagereward} to evaluate image quality. Therefore, DSH-Bench evaluates from the three aforementioned dimensions.

\textbf{Subject Preservation}\quad DreamBench++ utilizes GPT-4o for evaluation to improve alignment with human assessments. However, the GPT-4o-based method is prohibitively expensive. To address this limitation, we propose a novel metric---\textbf{Subject Identity Consistency Score} (SICS). As shown in \cref{fig:SCIS process}, we establish a scoring criterion for assessing subject preservation firstly. Five annotators label the collected image pairs according to the criterion. During the annotation process, each image pair is not only assigned a score but also accompanied by an explanation. Previous work~\cite{wei2022chain} has indicated that labeled data with explanatory reasoning can help models better understand the underlying logic and reasoning behind the labels. We then perform meticulous fine-tuning of the model using this annotated dataset. During fine-tuning, SICS leverages prompts to explicitly prioritize subject consistency rather than global semantics, mitigating background and style artifacts that commonly bias CLIP-based approaches and yielding closer alignment with the goals of subject-consistency evaluation. Although GPT-4o demonstrates outstanding performance across a wide range of tasks, it has not been specifically optimized for subject preservation evaluation. More details of the SICS metric can be found in Supplementary material (Sec. E.2).

\textbf{Prompt Following}\quad Prompt following primarily evaluates whether a model can generate images that accurately correspond to textual prompts. DreamBench++ has demonstrated that the CLIP-T score is highly consistent with human annotations. Therefore, we also adopt CLIP-T score as the evaluation metric for prompt following.

\textbf{Image Quality}\quad HPSv2~\cite{wu2023human} utilizes professionally annotated data to more accurately reflect human aesthetic preferences for generated images. Previous studies~\cite{sun2025ie} demonstrate that models optimized with HPSv2 achieve superior performance in image quality assessment compared to existing approaches. Therefore, we adopt HPSv2 for image quality evaluation.


\section{Experiment}
\subsection{Experiment Setup}
\label{sec:experiment setup}

\begin{figure*}[t]
\centering
\includegraphics[width=1.0\textwidth]{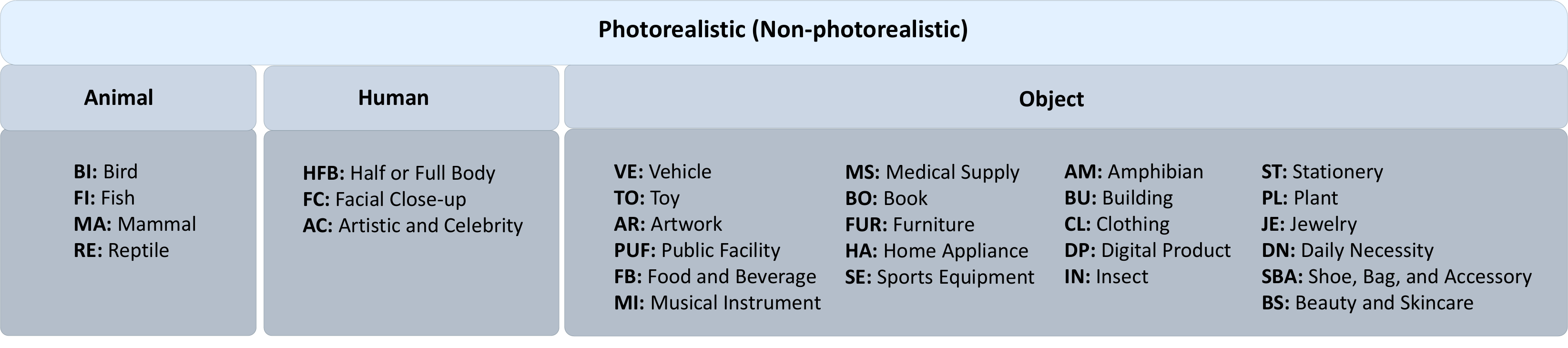}
\caption{\textbf{Category hierarchy of the dataset.} The top-level categories Photorealistic and Non-photorealistic share an identical set of sub-categories.}
\label{fig:cate_dist_intro_eccv}
\end{figure*}
\textbf{Implementation Details} \quad We conducted experiments on the following 19 models in total: 1) Textual Inversion(TI)~\cite{gal2023an}, 2) DreamBooth, 3) Custom Diffusion, 4) Hiper~\cite{han2023highly}, 5) NeTI~\cite{alaluf2023neural}, 6) BLIP-Diffusion~\cite{li2023blip}, 7) IP-Adapter~\cite{ye2023ip}, 8) MS-Diffusion~\cite{wang2024ms}, 9) Emu2~\cite{sun2024generative}, 10) OminiControl~\cite{tan2024ominicontrol}, 11) SSR-Encoder~\cite{zhang2024ssr}, 12) RealCustom++~\cite{mao2024realcustom++}, 13) OmniGen~\cite{xiao2024omnigen}, 14) $\lambda$-Eclipse~\cite{patel2024lambda}, 15) UNO~\cite{wu2025less}, 16) ACE++~\cite{mao2025ace++}, 17) DreamO~\cite{mou2025dreamo}, 18) FLUX.1 Kontext [dev]~\cite{batifol2025flux}, 19) Nano-Banana~\cite{nano-banana}. Our experiments are conducted using the official implementations to guarantee reliability and fairness. To demonstrate that our benchmark remains challenging for advanced closed-source models, we conducted experiments using Nano-Banana. More details can be found in supplementary material (Sec. E). 

\textbf{Dataset} \quad Our benchmark ultimately comprises 459 subject images and 5,508 prompts. These images are distributed across 58 distinct categories. Detailed information regarding category distribution is provided in \cref{fig:cate_dist_intro_eccv}.

\textbf{Human Annotation} \quad All annotation tasks, including labeling of the SICS training datasets, were conducted by the same five human annotators. We provide the annotators with detailed labeling guidelines and sufficient training to ensure they fully understand the subject-driven T2I generation task and could provide unbiased and discriminative scores. For additional details regarding the human annotation process, please see the supplementary material (Sec. E.4).

\setlength{\tabcolsep}{7.5pt}
\begin{table}[tb]
\caption{\label{table:SICS_res} The human alignment degree among different metrics, measured by \textbf{Kendall's $\tau$ value} and \textbf{Spearman correlation coefficient value}. H: Human, G: GPT-4o, D: DINO, Dv2: DINOv2, CB: CLIP-B, CL: CLIP-L, S: SICS. Bold font is used to denote the maximum value in a row.} 
\centering
\scalebox{0.6}{
\begin{tabular}{l|cccccc|cccccc}
\noalign{\smallskip}\toprule

\multirow{2}*{\textbf{Method}} & \multicolumn{6}{c|}{\textbf{Kendall$\uparrow$}} & \multicolumn{6}{c}{\textbf{Spearman$\uparrow$}} \\
\cline{2-13}
~& H-CB & H-CL & H-D & H-Dv2 & H-G & H-S & H-CB & H-CL & H-D & H-Dv2 & H-G & H-S \\
\hline  \noalign{\smallskip}

BLIP-Diffusion & 0.228 & 0.176 & 0.285 & 0.167 & \uline{0.354} & \textbf{0.531} & 0.285 & 0.215 & 0.350 & 0.206 & \uline{0.383} & \textbf{0.554}  \\
IP-Adapter & 0.294 & 0.296 & 0.258 & 0.290 & \uline{0.419} & \textbf{0.622} & 0.364 & 0.371 & 0.325 & 0.364 & \uline{0.459} & \textbf{0.657}  \\
MS-Diffusion & \uline{0.158}  &	0.090 & 0.116 & 0.122 & 0.119 & \textbf{0.178} & \textbf{0.194} & 0.109  &	0.144 &	0.156 &	0.131 &	\uline{0.189} \\
OminiControl   & 0.375 & 0.371 & 0.337 & 0.348 & \uline{0.650} & \textbf{0.713} & 0.490 & 0.486 & 0.441 & 0.453 & \uline{0.729} & \textbf{0.764}  \\
SSR-Encoder   & 0.264 & 0.338 & 0.295 & 0.348 & \uline{0.504} & \textbf{0.664} & 0.328 & 0.421 & 0.368 & 0.434 & \uline{0.549} & \textbf{0.697}  \\
ACE++ & 0.341 & 0.325 & 0.312 & 0.312 & \uline{0.425} & \textbf{0.524} & 0.298 & 0.352 & 0.323 & \uline{0.415} & 0.313 & \textbf{0.492}  \\
DreamO & 0.243 & 0.311 & 0.219 & 0.243 & \uline{0.321} & \textbf{0.361} & 0.240 & 0.291 & 0.291 & 0.311 & \uline{0.322} & \textbf{0.383}  \\
FLUX.1 Kontext [dev] & 0.346 & 0.313 & 0.269 & 0.340 & \uline{0.413} & \textbf{0.465} & 0.371 & 0.394 & \uline{0.430} & 0.361 & 0.297 & \textbf{0.496}  \\
UNO   & 0.249 & 0.218 & \uline{0.299} & 0.240 & 0.236 & \textbf{0.385} & 0.340 & 0.297 & \uline{0.390} & 0.312 & 0.268 & \textbf{0.426}  \\
RealCustom++   & 0.181 & 0.128 & 0.206 & 0.241 & \uline{0.291} & \textbf{0.464} & 0.229 & 0.162 & 0.266 & 0.303 & \uline{0.325} & \textbf{0.511}  \\
OmniGen  & 0.465 & 0.396 & 0.344 & 0.349 & \uline{0.617} & \textbf{0.621} & 0.579 & 0.497 & 0.440 & 0.456 & \textbf{0.697} & \uline{0.667}  \\
$\lambda$-Eclipse   & 0.143 & 0.233 & 0.084 & 0.103 & \uline{0.325} & \textbf{0.375} & 0.176 & 0.287 & 0.103 & 0.127 & \uline{0.352} & \textbf{0.393}  \\
Custom Diffusion   & 0.316 & 0.336 & 0.382 & 0.425 & \uline{0.487} & \textbf{0.642} & 0.388 & 0.409 & 0.470 & \uline{0.519} & 0.512 & \textbf{0.654}  \\
DreamBooth   & 0.639 & 0.591 & 0.537 & 0.429 & 0.647 & \textbf{0.692} & \uline{0.733} & 0.721 & 0.661 & 0.537 & 0.705 & \textbf{0.740}  \\
Textual Inversion  & 0.482 & 0.459 & 0.447 & 0.438 & \uline{0.541} & \textbf{0.568} & \uline{0.587} & 0.559 & 0.545 & 0.534 & 0.582 & \textbf{0.590}  \\
HiPer   & 0.338 & 0.387 & 0.351 & 0.404 & \uline{0.584} & \textbf{0.625} & 0.417 & 0.469 & 0.430 & 0.496 & \uline{0.629} & \textbf{0.655}  \\
NeTI   & 0.469 & 0.456 & 0.431 & 0.417 & \uline{0.617} & \textbf{0.728} & 0.573 & 0.561 & 0.529 & 0.512 & \uline{0.682} & \textbf{0.778}  \\
\noalign{\smallskip}\hline\noalign{\smallskip}
ALL   & 0.416 & 0.411 & 0.350 & 0.376 & \uline{0.619} & \textbf{0.677} & 0.529 & 0.522 & 0.451 & 0.483 & \uline{0.697} & \textbf{0.734} \\

\bottomrule

\end{tabular}
}

\end{table}
\setlength{\tabcolsep}{1.4pt}

\setlength{\tabcolsep}{8pt}
\begin{table}[t]

\caption{\label{table:overall} \textbf{Evaluation of Subject-driven T2I generation on open-source models.} DB: DreamBench, DB++: DreamBench++, HB: DSH-Bench. All scores are normalized to 0-1. Bold indicates \textbf{the minimum value in each row} for a given evaluation dimension. Experiments show that DSH-Bench is more difficult.}

\centering
\scalebox{0.7}{
\begin{tabular}{l|ccc|ccc|ccc}
\noalign{\smallskip}\toprule

\multirow{2}*{\textbf{Method}}&
\multicolumn{3}{c|}{\textbf{Subject Preservation}}&
\multicolumn{3}{c|}{\textbf{Prompt Following}}&
\multicolumn{3}{c}{\textbf{Image Quality}} \\
\cline{2-10}

 ~& 
DB & DB++ & HB & 
DB & DB++ & HB & 
DB & DB++ & HB      \\ 


\hline 
 \noalign{\smallskip}

BLIP-Diffusion &
0.229  & 0.216  & \textbf{0.204}  & 0.291  & 0.278  & \textbf{0.277}  & 0.267  & 0.254  & \textbf{0.223} \\

IP-Adapter & 
0.230  & 0.244  & \textbf{0.229}  & 0.321  & 0.318  & \textbf{0.315}  & 0.291  & 0.296  & \textbf{0.266}  \\

MS-Diffusion & 
\textbf{0.316}  & 0.346  & 0.352  & \textbf{0.332}  & 0.339  & 0.338  & 0.311  & 0.314  & \textbf{0.294}  \\

OminiControl & 
0.279  & 0.268  & \textbf{0.258}  & \textbf{0.325}  & 0.337  & 0.334  & 0.312  & 0.308  & \textbf{0.290}  \\

SSR-Encoder & 
0.231  & \textbf{0.202}  & \textbf{0.202}  & 0.290  & \textbf{0.287}  & 0.295  & 0.273  & 0.270  & \textbf{0.247}  \\

ACE++ & 
0.324  & 0.303  & \textbf{0.292}  & 0.321  & 0.316  & \textbf{0.304}  & 0.294  & 0.303  & \textbf{0.252}  \\

DreamO & 
0.412  & 0.396  & \textbf{0.391}  & 0.324  & 0.339  & \textbf{0.326}  & 0.314  & 0.308  & \textbf{0.283}  \\

FLUX.1 Kontext [dev] & 
0.445  & 0.432  & \textbf{0.424}  & 0.321  & 0.324  & \textbf{0.319}  & 0.273  & 0.270  & \textbf{0.288}  \\

UNO & 
\textbf{0.409}  & 0.410  & \textbf{0.409}  & \textbf{0.317}  & 0.322  & 0.323  & 0.304  & 0.297  & \textbf{0.278}  \\

Emu2 & 
0.360  & 0.343  & \textbf{0.341}  & \textbf{0.291}  & 0.309  & 0.304  & 0.272  & 0.278  & \textbf{0.260}  \\

RealCustom++ & 
0.377  & 0.380  & \textbf{0.375}  & \textbf{0.325}  & 0.329  & 0.332  & 0.316  & 0.314  & \textbf{0.298}  \\

\bottomrule

\end{tabular}
}

\end{table}
\setlength{\tabcolsep}{1.4pt}

\setlength{\tabcolsep}{13pt}
\begin{table}[tb]
\caption{\label{table:ldboard} \textbf{ DSH-Bench leaderboard.} The models are ranked by the final score $\bm{S_h}$.}
\centering
\scalebox{0.68}{
\begin{tabular}{llcccc}
\noalign{\smallskip}\toprule\noalign{\smallskip}

\multirow{2}*{\textbf{Method}}&
\multirow{2}*{\textbf{T2I Model}}&
\multicolumn{1}{c}{\textbf{Subject}}&
\multicolumn{1}{c}{\textbf{Prompt}}&
\multicolumn{1}{c}{\textbf{Image}} &
\multirow{2}*{\textbf{$\bm{S_h}$}$\uparrow$} \\

~& ~& 
\multicolumn{1}{c}{\textbf{Preservation}}&  
\multicolumn{1}{c}{\textbf{Following}}& 
\multicolumn{1}{c}{\textbf{Quality}}& 
~  \\ 
\noalign{\smallskip} \hline 
\noalign{\smallskip}

Nano-Banana&	-	&	\textbf{0.439} 	&	\uline{0.337} 	&	\textbf{0.302} 	&	\textbf{0.272} 	\\
FLUX.1 Kontext [dev]	&	FLUX.1 Kontext	&	\uline{0.424} 	&	0.319 	&	0.288 	&	\uline{0.256} 	\\
UNO	&	FLUX.1-dev	&	0.409 	&	0.323 	&	0.278 	&	0.252 	\\
DreamO	&	FLUX.1-dev	&	0.391 	&	0.326 	&	0.283 	&	0.251 	\\
RealCustom++	&	SDXL	&	0.375 	&	0.332 	&	\uline{0.294} 	&	0.251 	\\
MS-Diffusion	&	SDXL	&	0.352 	&	\textbf{0.338}	&	\uline{0.294} 	&	0.248 	\\
Emu2	&	SDXL	&	0.341 	&	0.304 	&	0.260 	&	0.228 	\\
OminiControl	&	FLUX.1-schnell	&	0.258 	&	0.334 	&	0.290 	&	0.218 	\\
ACE++	&	FLUX.1-dev	&	0.292 	&	0.304 	&	0.252 	& 0.214 	\\
IP-Adapter	&	SDXL	&	0.256 	&	0.292 	&	0.266 	&	0.199 	\\
$\lambda$-Eclipse	&	SDXL	&	0.229 	&	0.315 	&	0.242 	&	0.198 	\\
OmniGen	&	SD v1.5	&	0.202 	&	0.295 	&	0.265 	&	0.183 	\\
SSR-Encoder	&	SDXL	&	0.188 	&	0.322 	&	0.247 	&	0.181 	\\
NeTI	&	SD v1.4	&	0.192 	&	0.301 	&	0.234 	&	0.176 	\\
BLIP-Diffusion	&	SD v1.5	&	0.204 	&	0.277 	&	0.223 	&	0.174 	\\
DreamBooth	&	SD v1.5	&	0.158 	&	0.321 	&	0.245 	&	0.164 	\\
HiPer	&	SD v1.4	&	0.135 	&	0.318 	&	0.247 	&	0.151 	\\
Textual Inversion	&	SD v1.5	&	0.109 	&	0.299 	&	0.225 	&	0.129 	\\
Custom Diffusion	&	SD v1.4	&	0.062 	&	0.323 	&	0.240 	&	0.091 	\\

\bottomrule
\end{tabular} }
\end{table}

\setlength{\tabcolsep}{1.4pt}

\subsection{Main Results}
\label{sec:main results}

\textbf{SICS Results}\quad \cref{table:SICS_res} presents a rigorous study of human alignment using \textit{Kendall's $\tau$ value} (KDV) and \textit{Spearman correlation coefficient value} (SCV) (metric selection rationale in supplementary material (Sec. E.2)). Notably, the image pairs used for this human-alignment study form a held-out evaluation set of 1{,}600 pairs that is \textit{completely disjoint} from the 5{,}000 image-text pairs used to fine-tune SICS, ensuring that the reported correlations reflect genuine human alignment rather than fitting to the training data (annotation protocol detailed in supplementary material (Sec. E.4)). Our experimental results demonstrate that \textbf{SICS achieves superior alignment with human evaluations compared to existing methods}, showing consistently higher agreement across both correlation metrics in most experimental settings. Although SICS attains second-highest correlation scores in MS-Diffusion and OmniGen, it significantly outperforms GPT-4o (\textit{GPT-4o refers to the evaluation method used in DreamBench++}) by \textbf{9.37\%} (KDV) and \textbf{5.31\%} (SCV). This performance gap strongly suggests SICS's enhanced capability in modeling human evaluation. Notably, GPT-4o exhibits greater consistency with human evaluation than CLIP and DINO, aligning with DreamBench++ findings. Importantly, our proposed SICS metric surpasses all existing metrics in human judgment consistency.

\textbf{Quantitative \& Qualitative Results} \quad \cref{table:overall} shows overall evaluation results. The results show that: \textbf{i)} \textbf{DSH-Bench poses more significant challenges than existing benchmarks.} For subject preservation and image quality, the majority of methods consistently yield lower scores on DSH-Bench. The result can be attributed to the hierarchical taxonomy sampling method employed, which allows our dataset to more accurately represent the true data distribution. Moreover, it highlights that benchmarks derived from true distributions present greater challenges. \textbf{ii)} For prompt following, DreamBench yields slightly lower scores than DSH-Bench for certain methods. In DreamBench, prompts requiring attribute change constitute 22.7\%, which is higher than the 16.7\% observed in DSH-Bench. \cref{fig:stop_2} indicates that all methods exhibit relatively poor average performance on prompts involving attribute change. \textbf{iii)} \cref{table:ldboard} shows that there exists a trade-off between subject preservation and prompt following. We plot the Pareto frontier (see in supplementary material (Sec. D.1)) using the data presented in \cref{table:ldboard}. The primary objective is to identify a Pareto optimal solution that effectively balances the two objectives. \textit{Additional results and discussions are in supplementary material (Sec. D.2)}.

\textbf{Leaderboard}\quad In order to assess a model’s overall capability, we define the final score as:
\begin{equation}
\label{eq:final score}
\mathcal{S}_{\text{h}} = \frac{3}{\frac{\lambda}{\mathrm{SP}}+\frac{\gamma}{\mathrm{PF}}+\frac{\mu}{\mathrm{IQ}}}
\end{equation}
SP, PF, and IQ represent the scores for Subject Preservation, Prompt Following, and Image Quality, respectively. $\lambda,\gamma,\mu$ are the weights assigned to the importance of each corresponding dimension. In this study, we set $\lambda=1.5, \gamma=1.5, \mu=1$, as subject preservation and prompt following are of paramount importance in subject-driven T2I generation. The harmonic mean requires strong performance across all dimensions to yield a high overall score. We rank models by $S_{h}$ scores in \cref{table:ldboard}. Nano-Banana exhibits relatively strong overall performance. Among open-source models, FLUX.1 Kontext [dev] performs best.

\section{Analysis}
\label{sec:analysis}

\begin{figure*}[t]
\centering
\includegraphics[width=1.0\textwidth, trim=0mm 5mm 0mm 5mm]{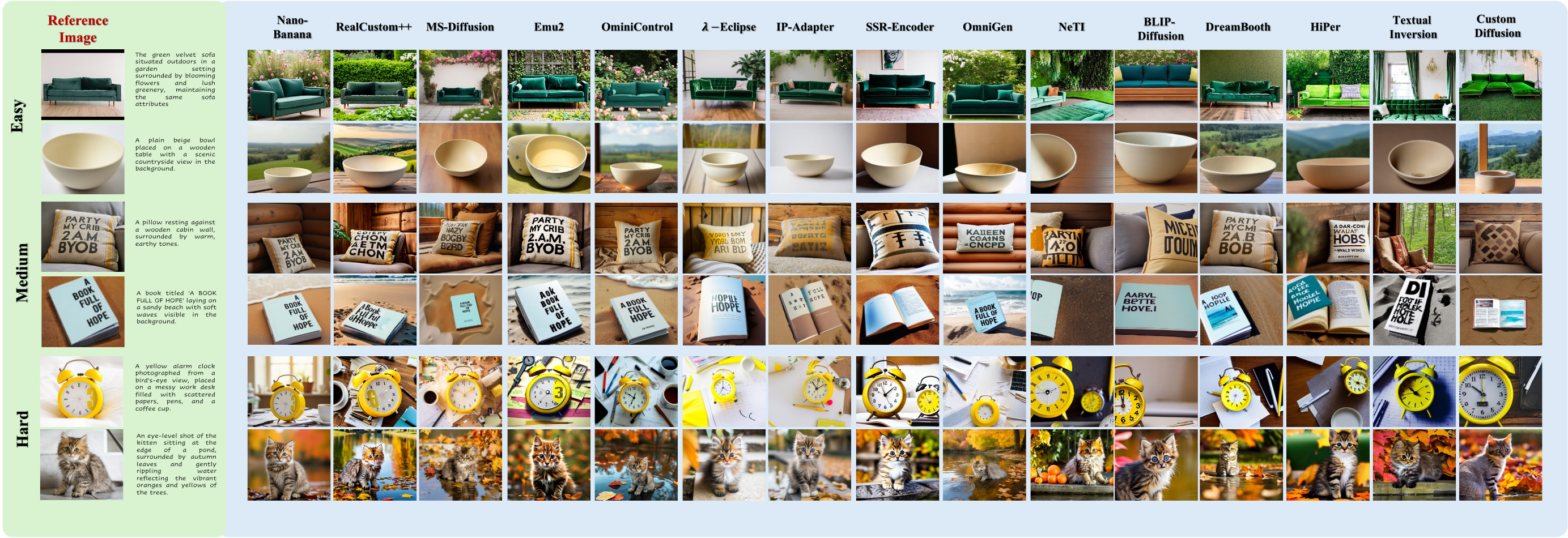}
\caption{Examples generated by methods listed in the leaderboard.}
\label{fig:ICLR_main_gen_exam_v1}
\end{figure*}

In this section, we conduct a detailed analysis of the performance of all methods based on the hierarchical category, subject difficulty level, and prompt scenario. \cref{fig:ICLR_main_gen_exam_v1} shows qualitative examples generated by the methods in the leaderboard.

\begin{figure}[tbp]
    \centering
    \begin{subfigure}{1.0\linewidth}
        \centering
        \includegraphics[width=\linewidth, trim=0mm 5mm 35.5mm 0]{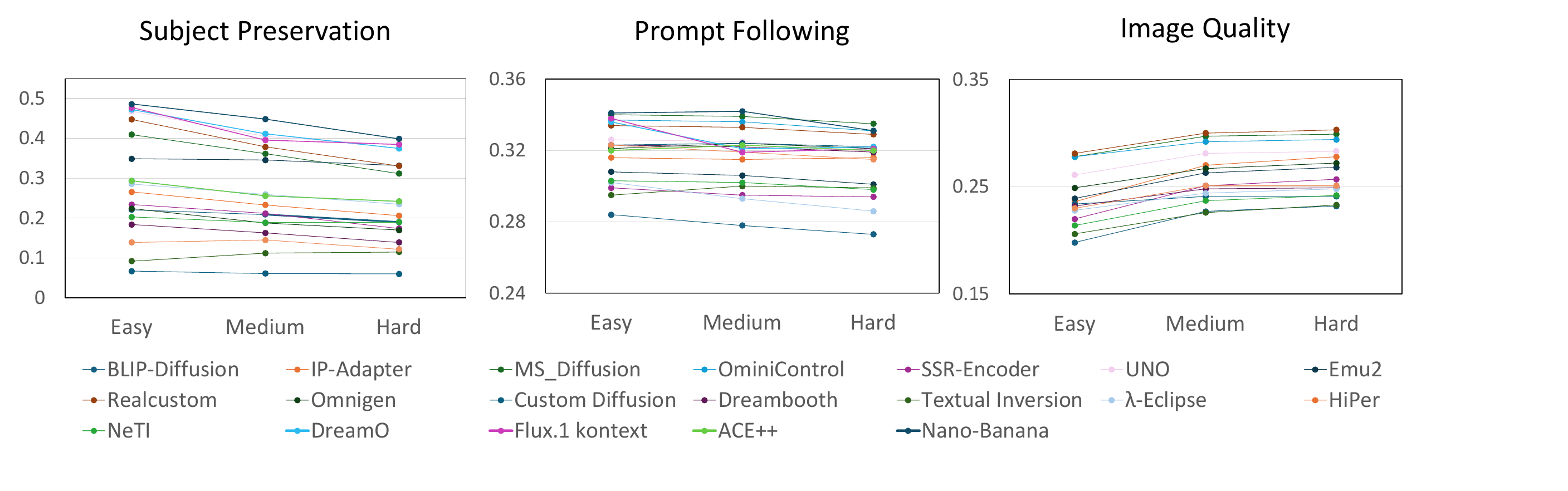}
        \caption{DSH-Bench scores across different subject difficulty level.}
        \label{fig:stop_1}
    \end{subfigure}
    \\
    \begin{subfigure}{1.0\linewidth}
        \centering
        \includegraphics[width=\linewidth, trim=3mm 0mm 0mm 0]{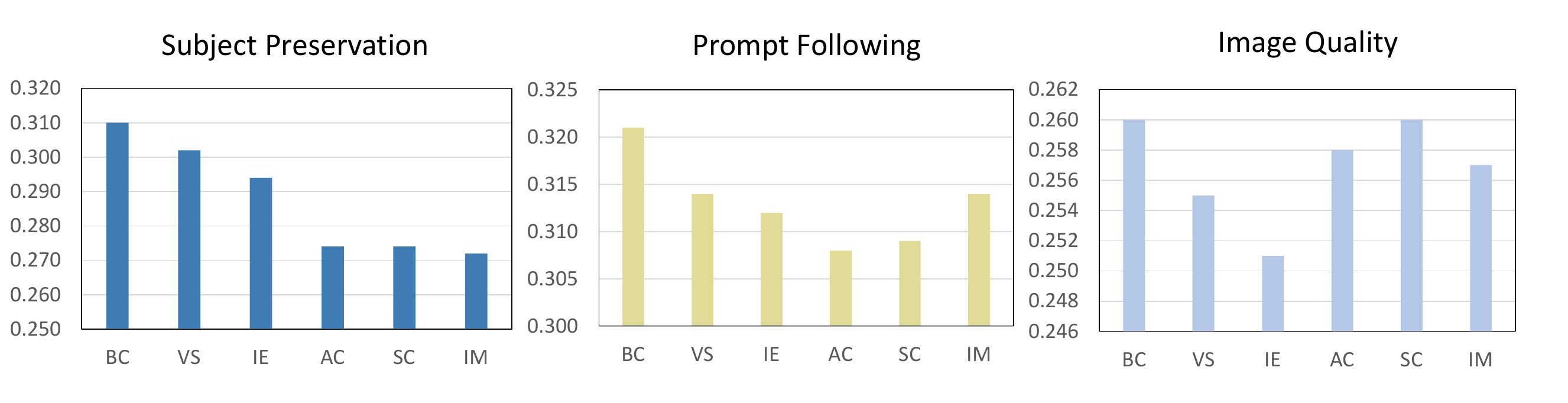}
        \caption{DSH-Bench scores across different prompt scenarios.}
        \label{fig:stop_2}
    \end{subfigure}
    \\
    \begin{subfigure}{1.0\linewidth}
        \centering
        \includegraphics[width=\linewidth, trim=0mm 0mm 20mm 0]{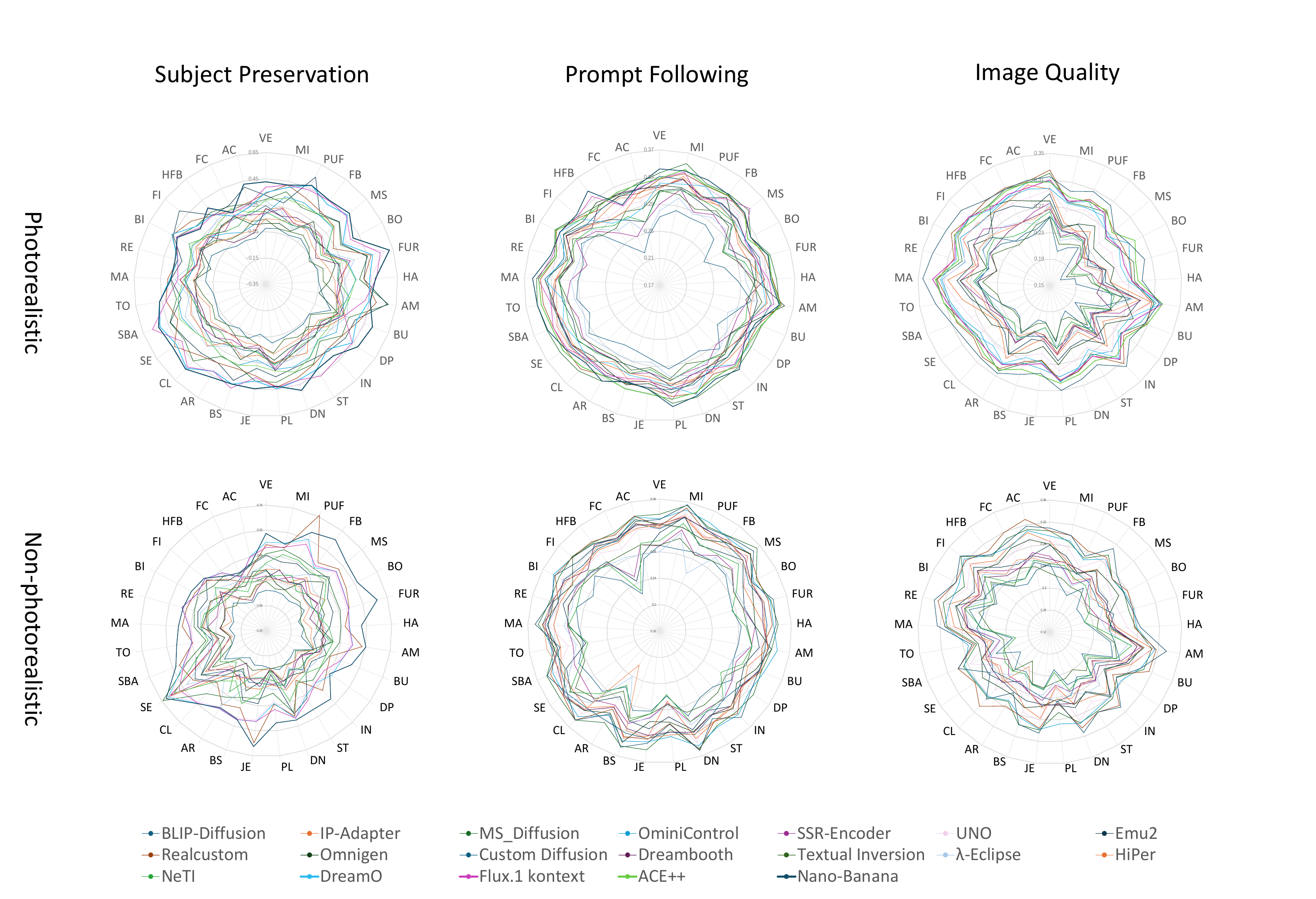}
        \caption{DSH-Bench scores across different categories.}
        \label{fig:stop_3}
    \end{subfigure}
    \caption{\textbf{Comparison for DSH-Bench scores across different evaluation dimensions.} Note that the categories are divided into two types: Photorealistic and Non-photorealistic. The specific metric values are provided in the supplementary material (Sec. D.2). Category definitions refer to supplementary material (Sec. E.3). Best viewed when zoomed in.}
    \label{fig:analysis}
\end{figure}




\textbf{A scientific and comprehensive subject image sampling method is necessary}\quad \cref{fig:stop_3} present the performance of various methods in the third-level categories. The results reveal that model robustness varies considerably among categories. For example, performance in category "\textit{Book}" (both photorealistic and non-photorealistic) is substantially lower. This disparity suggests that the absence of subject images from specific categories can lead to biased evaluation results, highlighting the importance of data diversity. Furthermore, \cref{fig:stop_3} also demonstrates that none of the current models perform well across all categories. We hypothesize that this may be related to the varying complexity of the subjects within different categories. A more detailed analysis of model performance in different categories can be found in supplementary material (Sec D.1).



\textbf{Current subject-driven T2I models exhibit performance degradation on hard level subjects}\quad As illustrated in \cref{fig:stop_1}, the model exhibits substantial variation in performance across different difficulty levels: 1) For subject preservation, there is a pronounced decline in performance as the difficulty of the subject images increases. The model achieves significantly better results on images classified as simple compared to those categorized as hard. This observation supports the validity of our image difficulty classification scheme. 2) As demonstrated in \cref{table:ldboard}, current closed-source models have achieved remarkable advancements in subject-driven T2I tasks, yielding highly competitive scores in our benchmark evaluation. Nevertheless, as illustrated in \cref{fig:stop_3}, the performance of Nano-Banana on challenging categories leaves ample room for improvement. Thus, our benchmark retains its formidable challenge. 3) For prompt following, \cref{fig:stop_1} shows that model capability is minimally influenced by the subject difficulty level. This could be explained by the fact that CLIP-T primarily emphasizes overall semantic information. Hence, as long as the generated image correctly represents the general category and overall shape, the evaluation score is unlikely to be substantially reduced, even if finer details are not perfectly captured. \textit{Given these findings, it is crucial to enhance models’ ability to encode and reconstruct complex subject details more effectively in future research}.




\textbf{The subject-driven T2I capability for different prompt scenarios is not robust}\quad \cref{fig:stop_2} shows the average performance of all models across six prompt scenarios. The results show that: 1) In BC, VS, and IE scenarios, the model's performance consistently declines across all evaluation dimensions. This trend suggests that scenario difficulty increases progressively from BC to IE. Notably, the finding that the IE scenario is more challenging than the BC scenario aligns with intuitive expectations. 2) For subject preservation, the model's average performance across the AC, SC, and IM scenarios remains relatively low. This could be because the generated subjects undergo partial modifications relative to the original subjects in these three scenarios. \textit{Given these findings, more emphasis should be placed on enhancing methods for IE prompt scenario. For instance, increasing the volume of training data tailored to these specific contexts.}

\section{Conclusion}
\label{conclusion}
This paper introduces a novel benchmark called DSH-Bench, designed specifically for subject-driven T2I generation. Key features include: 1) a hierarchical category system in image collection to ensure both the diversity and comprehensiveness of subject images; 2) an innovative classification scheme for categorizing subject difficulty levels and prompt scenarios to obtain valuable insights; and 3) a human-aligned and more efficient metric for subject preservation. The benchmark is publicly available at \url{https://github.com/sunriverhzy/DSH-Bench} to support the advancement in future research.

%
%
\bibliographystyle{splncs04}
\bibliography{main}

\end{document}